\documentclass[10pt,twocolumn,letterpaper]{article}

\usepackage{cvpr,times}

\usepackage{amsmath,amsfonts,bm}

\def\eqref#1{equation~\ref{#1}}

\def\1{\bm{1}}

\DeclareMathAlphabet{\mathsfit}{\encodingdefault}{\sfdefault}{m}{sl}
\SetMathAlphabet{\mathsfit}{bold}{\encodingdefault}{\sfdefault}{bx}{n}

\usepackage{url}
\usepackage{makecell}
\usepackage{floatrow}
\usepackage{epsfig}
\usepackage{graphicx}
\usepackage{amsmath}
\usepackage{amssymb}
\usepackage[titletoc,title]{appendix}
\usepackage{comment}
\usepackage{array}
\usepackage{multirow}
\usepackage{enumitem}
\usepackage{wrapfig}
\usepackage{graphicx}
\usepackage{lipsum}
\usepackage{tabularx}

\usepackage{natbib}

\usepackage{caption} 

\usepackage{hyperref}
\hypersetup{pagebackref=true,breaklinks=true,letterpaper=true,colorlinks,bookmarks=false}

\cvprfinalcopy 

\def\cvprPaperID{5879} 

\ifcvprfinal\fi

\newcommand{\AO}{\mathrm{G_A}}
\newcommand{\DA}{\mathrm{D_A}}
\newcommand{\DS}{\mathrm{D_S}}
\newcommand{\GRNN}{\mathrm{GRNN}}
\newcommand{\SO}{\mathrm{G_S}}

\newcommand{\F}{\textbf{F}} %

\newcommand{\M}{\textbf{M}} %
\title{Embodied Language Grounding with 3D Visual Feature Representations}

\author{Mihir Prabhudesai\thanks{Equal contribution},
Hsiao-Yu Fish Tung\footnotemark[1]   ,
Syed Ashar Javed\footnotemark[1], \\
Maximilian Sieb\thanks{Work done  while in Carnegie Mellon University. }    , Adam W. Harley, Katerina Fragkiadaki \\
\texttt{\{mprabhud,htung,sajaved,msieb,aharley,katef\}@cs.cmu.edu} \\
Carnegie Mellon University \\
}

\begin{document}
\maketitle


\begin{abstract}
We  propose 
associating language utterances to 3D visual abstractions of the scene they describe. 
The 3D visual abstractions are encoded as 3-dimensional visual feature maps. 
We infer these 3D visual scene feature maps from RGB images of the scene via view prediction: 
when the generated 3D scene feature map is neurally projected from a camera viewpoint, it should  match the corresponding RGB image. 
We present generative models that  condition on the dependency tree of an utterance and generate a corresponding visual 3D feature map as well as reason about its plausibility, and detector models that condition on both the dependency tree of an utterance and a related image and localize the  object referents in the   3D feature map inferred from the  image.  
Our model outperforms  models of language and vision that associate language with 2D CNN activations or 2D images by a large margin in a variety of tasks, such as, 
 classifying plausibility of utterances, detecting referential expressions, and supplying rewards for trajectory optimization of object placement policies from language instructions.  
 We perform numerous ablations and show the improved performance of our detectors is due to its  better  generalization across   camera viewpoints and lack of object interferences in the inferred 3D feature space, and the improved performance of our generators is due to their ability to spatially reason about objects and their configurations in 3D when mapping from language to scenes.

\end{abstract}

\section{Introduction} \label{sec:intro}

Consider the utterance \textit{``the tomato is to the left of the pot"}. Humans can answer numerous questions about the situation  described such as, \textit{``is the pot larger than the tomato?", ``can we move to a viewpoint from which the tomato is completely hidden behind the pot?", ``can we have an object that is both to the left of the tomato and to the right of the pot?"}, and so on. 
How can we learn computational models  that would permit a machine to carry out similar types of reasoning? One possibility is to treat the task as text comprehension \cite{DBLP:journals/corr/WestonCB14,NIPS2015_5945,DBLP:journals/corr/KadlecSBK16,DBLP:journals/corr/DhingraLCS16} and  train machine learning models using supervision from utterances accompanied with  question answer pairs. However,  information needed for answering the questions is not contained in the utterance itself;  training a model to carry out predictions in absence of the relevant information would  lead to overfitting. Associating utterances with RGB images that depict the scene described in the utterance, and 
using both images and utterances for answering questions, provides more world context and has been shown to be helpful. Consider though that information about object size, object extent, occlusion relationships, free space and so on, are only indirectly present in an RGB image, while they are readily available given a 3D representation of the scene the image depicts. Though it would take many training examples to learn whether a spoon can be placed in between the tomato and the pot on the table, in 3D this experiment can be  imagined easily, simply by considering whether the 3D model of the spoon can  fit in the free space between the tomato and the pot. 
Humans are experts in inverting camera projection and inferring an approximate 3D scene given an RGB image \cite{Olshausen2013PerceptionAA}. 
This paper builds upon  inverse graphics neural architectures for providing the 3D visual representations to associate language,  with the hope to inject spatial reasoning capabilities into  architectures for language understanding.

\begin{figure*}[t!]
    \centering
    \includegraphics[width=\textwidth]{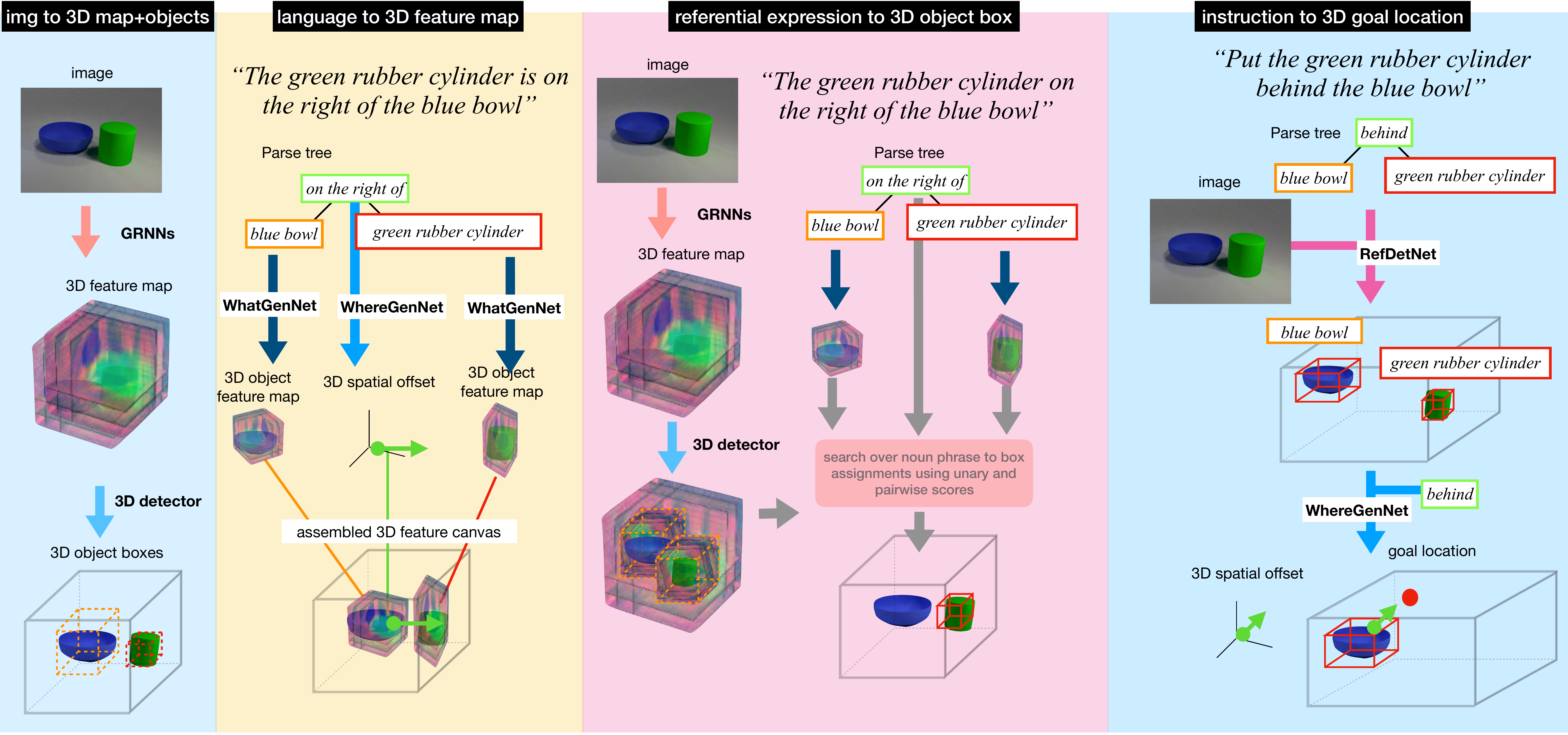}
    \caption{ \textbf{Embodied language grounding with 3D visual feature representations.}
    Our model associates utterances with 3D scene feature representations. 
    We map  RGB images to 3D scene feature representations and 3D object boxes of the objects present, building upon the method of Tung et al.~\cite{commonsense} (column 1). We map an utterance and its dependency  tree to object-centric 3D feature maps and cross-object relative 3D offsets using stochastic generative networks (column 2).  We map a referential expression to the 3D box of the object referent (column 3). 
    Last, given a placement instruction, we localize the referents in 3D in the scene and infer the desired 3D location for the object to be manipulated (column 4). We use the predicted location to  supply rewards for trajectory optimization of placement policies.
     }
    \label{fig:intro}
\end{figure*}

\textbf{We propose  associating  language utterances to space-aware 3D visual feature representations of the scene they describe}.  
We infer such 3D scene representations from RGB images of the   scene. 
Though inferring 3D scene representations from RGB images, a.k.a. inverse graphics, is known to be a  difficult problem \cite{DBLP:journals/corr/KulkarniWKT15,Romaszko_2017_ICCV,DBLP:journals/corr/TungHSF17}, we build upon recent advances in computer vision \cite{commonsense} that consider inferring from images a learnable 3D scene feature representation in place of explicit 3D representations such as meshes, pointclouds or binary voxel occupancies pursued in previous inverse graphics research \cite{DBLP:journals/corr/KulkarniWKT15,Romaszko_2017_ICCV,DBLP:journals/corr/TungHSF17}. These learnable 3D scene feature maps emerge in a self-supervised manner by optimizing for view prediction in neural architectures with geometry-aware 3D  representation bottlenecks \cite{commonsense}. 
After training, these architectures learn to map RGB video streams or single RGB images to complete 3D feature maps of the scene they depict, inpainting details that were occluded or missing from the 2D image input. 
\textbf{The contribution of our work is to use such 3D feature representations for language understanding and spatial reasoning.} We train modular generative networks that condition on the dependency tree of the utterance and predict a 3D feature map of the  scene the utterance describes. They do so by predicting the appearance and relative 3D location of  objects, and updating a 3D feature workspace, as shown in Figure \ref{fig:intro}, 2nd column. We further train modular discriminative networks that condition on a referential expression and detect the object being referred to,
by scoring object appearances and cross-object spatial arrangements, respectively, as shown  in  Figure \ref{fig:intro}, 3rd column. 
We call our model \textit{embodied} since training the 2D image to 3D feature mapping requires self-supervision by a mobile agent that moves around in the 3D world and collects posed images. 

We demonstrate the benefits of associating language to 3D visual feature scene representations in three basic language understanding tasks: 

\textbf{(1) Affordability reasoning. }
Our model can classify affordable (plausible) and unaffordable (implausible) spatial expressions. For example,  \textit{``A to the left of B, B to the left of C, C to the right of A"}  describes a plausible configuration, while \textit{``A to the left of B, B to the left of C, C to the left of A"} describes a non-plausible scene configuration, where A, B, C any object mentions. Our model reasons about plausibility of object arrangements in the inferred 3D feature map, where free space and object 3D intersection can  easily be learned/evaluated, as opposed to 2D image space.

\textbf{(2) Referential expression detection. } 
Given a referential spatial  expression, e.g., \textit{``the blue sphere behind the yellow cube"}, and an RGB image, our model outputs the 3D object bounding box of the referent in the inferred 3D feature map,  as shown in Figure \ref{fig:intro} 3rd column. 
Our 3D referential detector generalizes across  camera viewpoints better than existing state-of-the-art 2D referential detectors  \cite{modularreferential} thanks to the view invariant 3D feature representation.   

\textbf{(3) Instruction following. } 
Given an object placement instruction, e.g., \textit{``put the cube behind the book"}, our referential 3D object detector identifies the object to be manipulated, and our generative network predicts  
its desired 3D goal  location,  as shown in Figure \ref{fig:intro} 4th column. We use the predicted 3D goal  location  in trajectory optimization of object placement policies. We empirically show that our model successfully executes natural language instructions. 

In each task we compare against existing state-of-the-art models: the language-to-image generation model of Deng et al.~\cite{NIPS2018_7658} and the 2D referential object detector of Hu et al.~\cite{modularreferential}, which we adapt to have same input as our model. Our model outperforms the baselines by a large margin in each of the three tasks.  We further show strong generalization of natural language learned concepts from the simulation to the real world, thanks to the what-where decomposition employed in our generative and detection networks, where spatial expression detectors only use 3D spatial  information, as opposed to object appearance and generalize to drastically different looking scenes without any further annotations.  
Our model's  improved performance  is attributed to i) its improved generalization across camera placements thanks to the  viewpoint invariant 3D feature representations, and ii) its improved performance on free-space inference and plausible object placement in 3D over 2D. 
Many physical properties can be trivially evaluated in 3D while they need to be learned through a large number of training examples in 2D, with questionable generalization across viewpoints. 3D object intersection is one such property, which is useful for reasoning about plausible object arrangements. 

\section{Related  Work} \label{sec:related}

Learning and representing common sense world knowledge for language understanding is a major open research question.   
Researchers have considered grounding natural language on visual cues as a means of injecting visual common sense  to natural language understanding \cite{rohrbach2013translating,fang2015captions,rohrbach2013translating,fang2015captions,antol2015vqa,devlin2015language,andreas2016learning,rohrbach2012database,rohrbach2017generating,rohrbach2015dataset,karpathy2015deep,DBLP:journals/corr/YangHGDS15,donahue2015long,NIPS2018_7658}. For example visual question answering is a task that  has attracted a lot of attention and whose  performance has been steadily improving over the years \cite{DBLP:journals/corr/abs-1902-05660}. Yet, 
there is vast knowledge regarding basic physics and mechanics that current  vision and language models  miss,  as explained in Vedantam et al.~\cite{vedantamLinICCV15}. 
For example, existing models 
 cannot infer whether \textit{``the mug inside the pen"} or \textit{``the pen inside the mug"} is more plausible, whether \textit{``A in front of B, B in front of C, C in front of A"} is realisable, whether the mug  continues to exist if the camera changes viewpoint, and so on.  
 It is further unclear what supervision is necessary for such reasoning ability to emerge in current model architectures.

\section{Language grounding on 3D 
visual feature representations}
\label{sec:model}

We consider a dataset of 3D static scenes annotated with corresponding  language descriptions and their dependency trees, as well as  a reference camera viewpoint.  We further assume access at training time to  3D object bounding boxes and correspondences between 3D object boxes and noun phrases in the  language dependency trees. The language utterances we use describe object spatial arrangements and are programmatically generated, similar to their dependency trees,  using the method described in Johnson et al.~\cite{DBLP:journals/corr/JohnsonHMFZG16}. 
We infer  3D feature maps of the world scenes from RGB images using  Geometry-aware Recurrent Neural Nets (GRNNs) of Tung et al.~\cite{commonsense}, which we describe for completeness in Section \ref{sec:GRNN}. GRNNs  learn to map  2D image streams to 3D visual feature maps while optimizing for view prediction,  without any language supervision. 
 In Section \ref{sec:simsem}, we describe our proposed generative networks that condition on the dependency tree of a  language utterance and generate  an object-factorized 3D feature map of the scene the utterance depicts. 
 In Section \ref{sec:refdet}, we describe  discriminative networks  that condition on the dependency tree of a  language utterance  and the inferred 3D feature map from the RGB image and localize the  object being referred to in 3D. In Section \ref{sec:instr}, we show how our generative and discriminative networks of Sections \ref{sec:simsem} and \ref{sec:refdet} can be used to follow object placement instructions. 

\subsection{Inverse graphics  with Geometry-aware Recurrent Neural Nets (GRNNs)}
\label{sec:GRNN}
GRNNs learn to map  an RGB or RGB-D (color and depth) image or image sequence that depicts a static 3D world scene to a 3D feature map of the scene in an end-to-end differentiable manner while optimizing for view prediction: the inferred 3D feature maps, when projected from designated camera viewpoints, are neurally decoded to  2D RGB images and the weights of the neural architecture are trained to minimize RGB distance of the predicted image from the corresponding ground-truth RGB image view. We will denote the inferred 3D feature map  as $\M \in \mathbb{R}^{W \times H \times D \times C}$---where $W,H,D,C$ stand for width,  height,  depth and  number of feature channels, respectively. 
 Every $(x, y, z)$ grid location in the 3D feature map $\M$ holds a 1-dimensional feature vector  that  describes the semantic and geometric properties of a corresponding 3D physical location in the 3D world scene.   The  map is  updated with each new video frame while cancelling camera motion, 
 so that information from 2D pixels that correspond to the same 3D physical point end-up nearby in the map. 
 At training time, we assume a mobile agent that  moves around in a 3D world scene and sees it from multiple camera viewpoints, in order to provide ``labels" for view prediction to GRNNs. 
     Upon training, GRNNs can map an RGB or RGB-D image sequence or single image to a complete 3D feature map of the scene it depicts, i.e., it learns to \textit{imagine} the missing or occluded information; we denote this 2D-to-3D mapping as $\M=\GRNN(I)$ for an input RGB or RGB-D image $I$.
     
    \textbf{3D object proposals.} 
Given images with annotated 3D object boxes, the work of Tung et al.~\cite{commonsense}  trained GRNNs for 3D object detection by learning a neural module that takes as input the 3D feature map $\M$ inferred from the input image and outputs 3D bounding boxes and binary 3D voxel occupancies (3D segmentations) for the objects present in the map.   Their work essentially  adapted  
 the state-of-the-art 2D object detector Mask-RCNN \cite{maskrcnn} to have 3D input and output instead of 2D. We use the same architecture for our category-agnostic 3D region proposal network (3D RPN)  in Section \ref{sec:refdet}.  
For further details on GRNNs, please read Tung et al.~\cite{commonsense}.

\subsection{Language-conditioned 3D visual imagination} 
\label{sec:simsem}
\begin{figure}[t!]
    \centering
    \includegraphics[width=\textwidth]{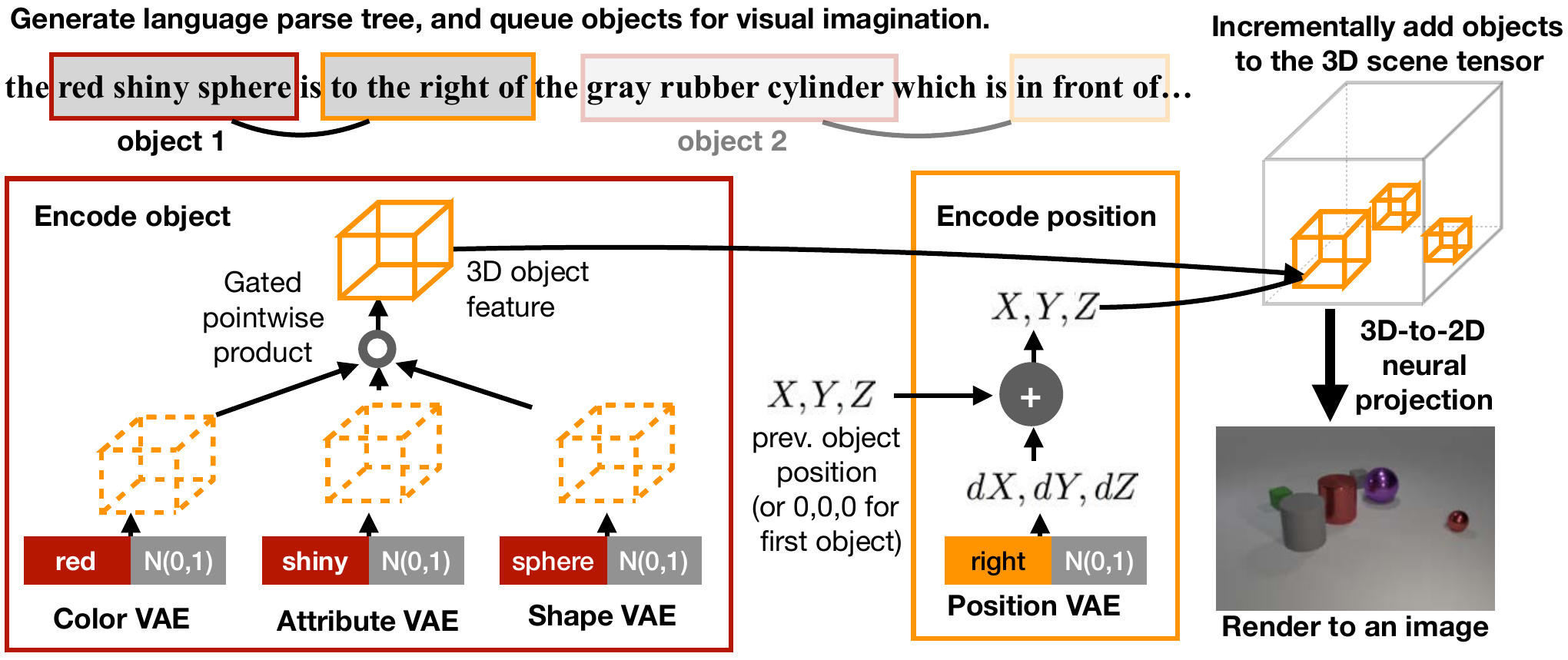}
    \caption{\textbf{Mapping  language utterances to object-centric appearance tensors and cross-object 3D spatial offsets} using conditional \textit{what}-\textit{where} generative networks.
    }
    \label{fig:VAE}
\end{figure}

We train generative networks to map  language utterances to 3D feature maps of the scene they describe. 
 They do so using a compositional generation process that conditions on the dependency tree of the utterance (assumed given) and generates one object at a time, predicting its appearance and location using two separate stochastic neural modules, \textit{what} and \textit{where}, as shown in Figure \ref{fig:VAE}.  

The {\it what} generation module $\AO(p, z; \phi)$ is a  stochastic generative network of object-centric appearance that given a noun phrase $p$ learns to map the word embeddings of each adjective and noun and a random vector of sampled Gaussian noise $z \in \mathbb{R}^{50} \sim \mathcal{N}(0,I)$ to a corresponding fixed size 3D feature tensor $ \bar{\M}^o \in \mathbb{R}^{w \times h \times d \times c}$ and a size vector $s^o \in \mathbb{R}^3$ that describes the width, height, and depth for the tensor.  We resize the 3D feature tensor $\bar{\M}^o$ to have the predicted size $s^o$ and obtain $\M^o=\mathrm{Resize}(\bar{\M}^o,s^o)$. 
We   
use a gated mixture of experts \cite{45929} layer---a gated version of point-wise multiplication---to aggregate outputs from different adjectives and nouns,  as shown in Figure \ref{fig:VAE}.

The {\it where} generation module $\SO(s, z, \psi)$ is a stochastic generative network of cross-object 3D offsets that learns to map  the one-hot encoding of a spatial expression $s$, e.g., \textit{``in front of"}, and  a random vector of sampled Gaussian noise $z \in \mathbb{R}^{50} \sim \mathcal{N}(0,I)$ to a relative 3D spatial offset $d\mathbf{X}^{(i,j)}=(dX,dY,dZ) \in \mathbb{R}^{3}$ between the corresponding objects. Let $b^o_i$
 denote the 3D spatial coordinates of the corners of a generated object.

Our complete generative network conditions on the dependency parse tree of the utterance and  adds one  3D object tensor $\M^{o}_i, i=1...K$ at a time to  a 3-dimensional feature canvas according to their predicted  3D locations, where $K$ is the number of noun phrases in the dependency tree: 
        $\M^g = \sum_{i=1}^K \mathrm{DRAW}(\M^{o}_i, \mathbf{X}^o_i), $
where $\mathrm{DRAW}$ denotes the operation of adding a 3D feature tensor to a 3D location. The 3D location $\mathbf{X}^1$ of the first object is chosen arbitrarily, and the locations of the rest of the object are based on the predicted cross-object offsets: $\mathbf{X}^o_2=\mathbf{X}^o_1+d\mathbf{X}^{(2,1)}$.  
 If two added objects intersect in 3D, i.e., the intersection over union of the 3D object bounding boxes is above a cross-validated threshold of 0.1, $\mathrm{IoU}(b^o_i,b^o_j)>0.1$,  we re-sample object locations until we find a scene configuration where  objects do not 3D intersect, or until we reach a maximum number of samples---in which case we infer that the utterance is impossible to realize.   By exploiting the constraint of non 3D intersection in the 3D feature space, our model can both generalize to longer parse trees than those seen at training time---by re-sampling until all spatial constraints are satisfied---as well as infer the plausibility of utterances, as we validate empirically in Section \ref{sec:afford}. 
 In 3D,  non-physically plausible object intersection is  easy to distinguish from physically plausible object occlusion, something that is not easy to infer with 2D object coordinates, as we show empirically in Section \ref{sec:afford}. 
 Given the 3D coordinates of two 3D bounding boxes, our model detects whether there exists 3D object interpenetration by simply thresholding the computed 3D intersection over union.
 
 
We train our stochastic generative networks using conditional variational autoencoders. We detail the inference networks in Section 1 of the supplementary file.


\subsection{Detecting referential expressions in 3D} \label{sec:refdet}

\begin{figure}[t!]
    \centering
    \includegraphics[width=0.9\textwidth]{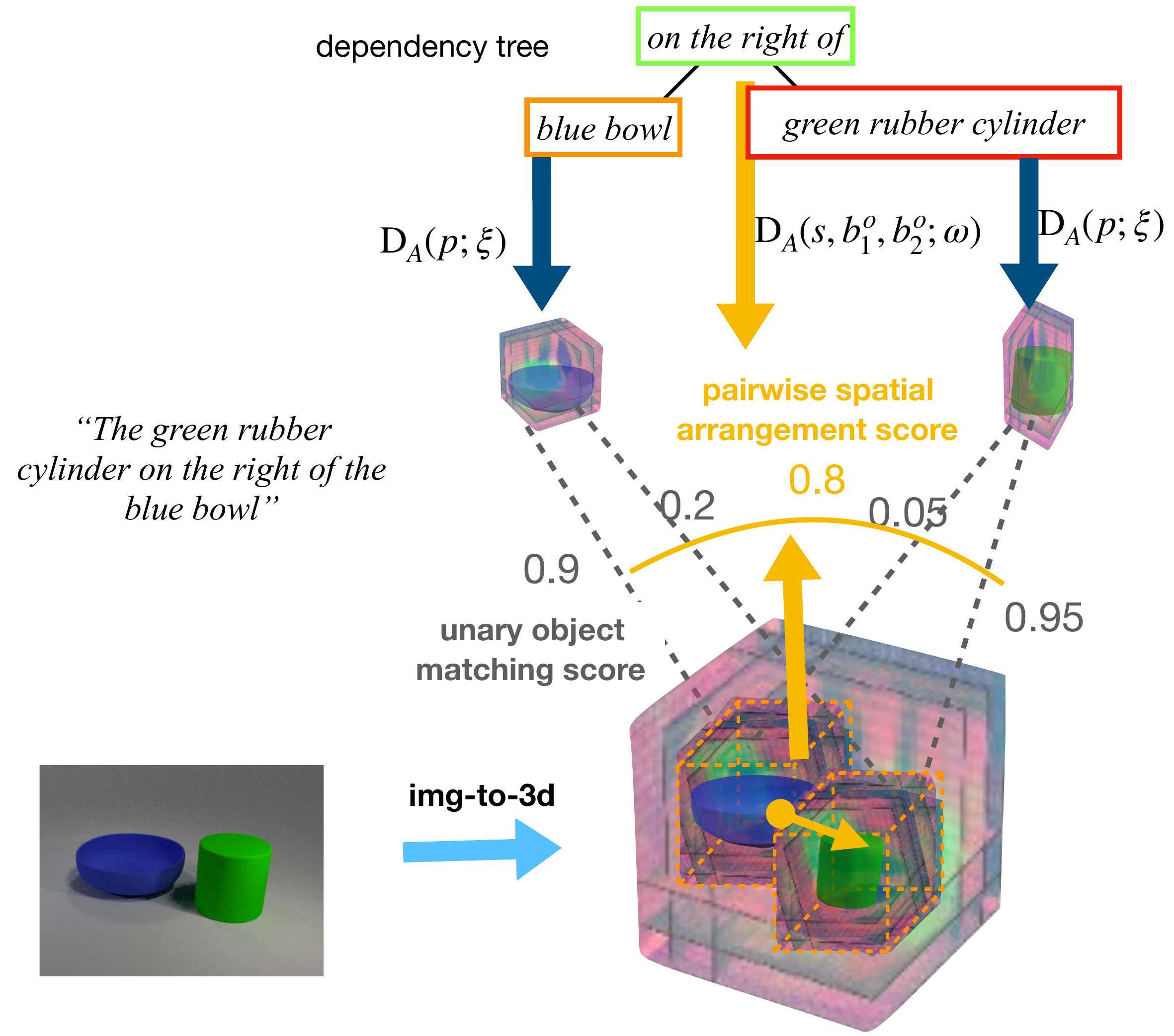}
    \caption{\textbf{3D referential object detection.} We score possible assignments of noun phrases to the detected 3D objects based on their
    appearance and pairwise spatial relations.
    }
    \label{fig:refdet}
\end{figure}

We train discriminative networks to map   spatial referential expressions, e.g., \textit{``the blue cube to the right of the yellow sphere behind the green cylinder"}, and related RGBD images, to the 3D bounding box of the objects the expressions refer to. 
Our model uses a compositional detection module conditioned on the dependency  tree of the referential expression (assumed given). The compositional detection module has two main components: (1) an object appearance matching function that predicts a 3D appearance detector template for each noun phrase and uses the template to compute an object appearance matching score, and (2) a 3D spatial classifier for each spatial expression that computes a spatial compatibility score. We detail these components below. The compositional structure of our detector is necessary to  handle referential expressions of arbitrary length.  
Our detector is comprised of a \textit{what} detection module and a \textit{where} detection module, as shown in Figure \ref{fig:refdet}. The \textit{what}  module $\DA(p;\xi)$ is a neural network that  given a noun phrase $p$ learns to map the word embeddings of each adjective and noun to a corresponding fixed-size 3D feature tensor $f=\DA(p;\xi) \in \mathbb{R}^{W \times H \times D \times C}$, we used $W=H=D=16$ and $C=32$. Our \textit{what} detection module is essentially a deterministic alternative of the \textit{what} generative stochastic network of Section \ref{sec:simsem}. 
The object appearance score is obtained by computing the inner product between  the detection template $\DA(p;\xi)$  and the cropped object 3D feature map $\F=\mathrm{CropAndResize}(\M,b^o)$, where $\M=\GRNN(I)$ and $b^o$ the 3D box of the object.  We feed the output of the inner product to a sigmoid activation layer. 

The \textit{where} detection module $\DS(s,b^o_1,b^o_2;\omega)$   takes as input  the 3D box coordinates of the hypothesized pair of objects under consideration, and  the one-hot encoding of the spatial utterance $s$ (e.g., \textit{``in front of", ``behind"}), and scores whether the two-object configuration matches the spatial expression.

We train both the \textit{what} and \textit{where} detection modules in a supervised way. During training, we use  ground-truth associations of noun phrases $p$ to 3D object boxes in the image  for positive examples, and random crops or other objects as negative examples. 
For cropping, we use ground-truth 3D object boxes   at training time and detected 3D object box proposals  from the 3D region proposal network (RPN) of Section \ref{sec:GRNN} at test time.  

Having trained  our \textit{what} and \textit{where} detector modules, and given the dependency parse tree of an utterance and a set of bottom up  3D object proposals, we exhaustively search over assignments of noun phrases to detected 3D objects in the scene. We only keep noun phrase to 3D box assignments if their unary matching score is above a cross-validated threshold of 0.4. Then, we simply pick the assignment of noun phrases to 3D boxes  with the highest product of unary and pairwise scores. 
Our 3D referential detector resembles previous 2D referential detectors \cite{modularreferential,Cirik2018UsingST}, but operates in 3D appearance features and spatial arrangements, instead of 2D.

\subsection{Instruction following} \label{sec:instr}
Humans use natural language to program fellow humans  e.g., \textit{``put the orange inside the wooden bowl, please''}. 
Programming robotic agents in a similar manner is desirable since it would allow non-experts to also program robots.
While most current policy learning methods use manually coded reward functions in simulation or instrumented environments to train policies, here we propose to use visual detectors of natural language expressions \cite{reward}, such as \textit{``orange inside the wooden basket,''} to automatically monitor an agent's progress towards achieving the desired goal and supply rewards accordingly. 

We use the language-conditioned generative and detection models proposed in Section \ref{sec:simsem} and \ref{sec:refdet} to obtain  a reliable perceptual reward detector for object placement instructions with the following steps, as shown in Figure \ref{fig:intro} 4th column: 
  \textbf{(1)}  We localize in 3D all  objects mentioned in the instruction using the aforementioned 3D referential  detectors. 
    \textbf{(2)}  We predict the desired 3D goal  location  for the object to be manipulated $\mathbf{x}^o_{goal}$ using our stochastic spatial arrangement generative network $\SO(s,z;\psi))$. 
    \textbf{(3)}  We compute per time step costs  being  proportional to the Euclidean distance of  the current 3D location  of the object $\mathbf{x}^o_t$  and end-effector 3D location $\mathbf{x}^e_t$ assumed known from forward dynamics, and  the desired 3D goal object location $\mathbf{x}^o_{goal}$ and end-effector 3D location   $\mathbf{x}^e_{goal}$: $\mathcal{C}_t=\|\mathbf{x}_t-\mathbf{x}_{goal}\|^2_2$, where $\mathbf{x}_t=[\mathbf{x}^o_t;\mathbf{x}^e_t]$ is the concatenation of object and end-effector state  at time step $t$ and $\mathbf{x}_{goal}=[\mathbf{x}^o_{goal};\mathbf{x}^e_{goal}]$.  We formulate this as a reinforcement learning problem, where at each time step the cost is given by  $c_t=\|\mathbf{x}_t-\mathbf{x}_{goal} \|_2$. 
We use i-LQR (iterative Linear Quadratic Regulator) \cite{Tassa2014ControllimitedDD} to minimize the cost function $\sum_{t=1}^T \mathcal{C}_t $. 
I-LQR learns a  time-dependent policy $\pi_t(\mathbf{u}|\mathbf{x};\theta)=\mathcal{N}(\mathbf{K}_t\mathbf{x}_t+\mathbf{k}_t, \mathbf{\Sigma}_t)$, where the time-dependent control gains are learned by model-based updates, where the dynamical model $p(\mathbf{x}_t|,\mathbf{x}_{t-1},\mathbf{u}_t)$ of the \textit{a priori} unknown dynamics is learned during training time. 
The actions $\mathbf{u}$ are defined as the changes in the robot end-effector's 3D position, and orientation about the vertical axis, giving a 4-dimensional action space.

We show in Section \ref{sec:instrfolexp} that our method successfully trains multiple language-conditioned policies. In comparison, 2D desired goal locations generated by 2D baselines \cite{reward} often fail to do so.

\section{Experiments}\label{sec:experiments}

 \begin{figure}[t!]
    \centering
    \includegraphics[width=1.0\textwidth]{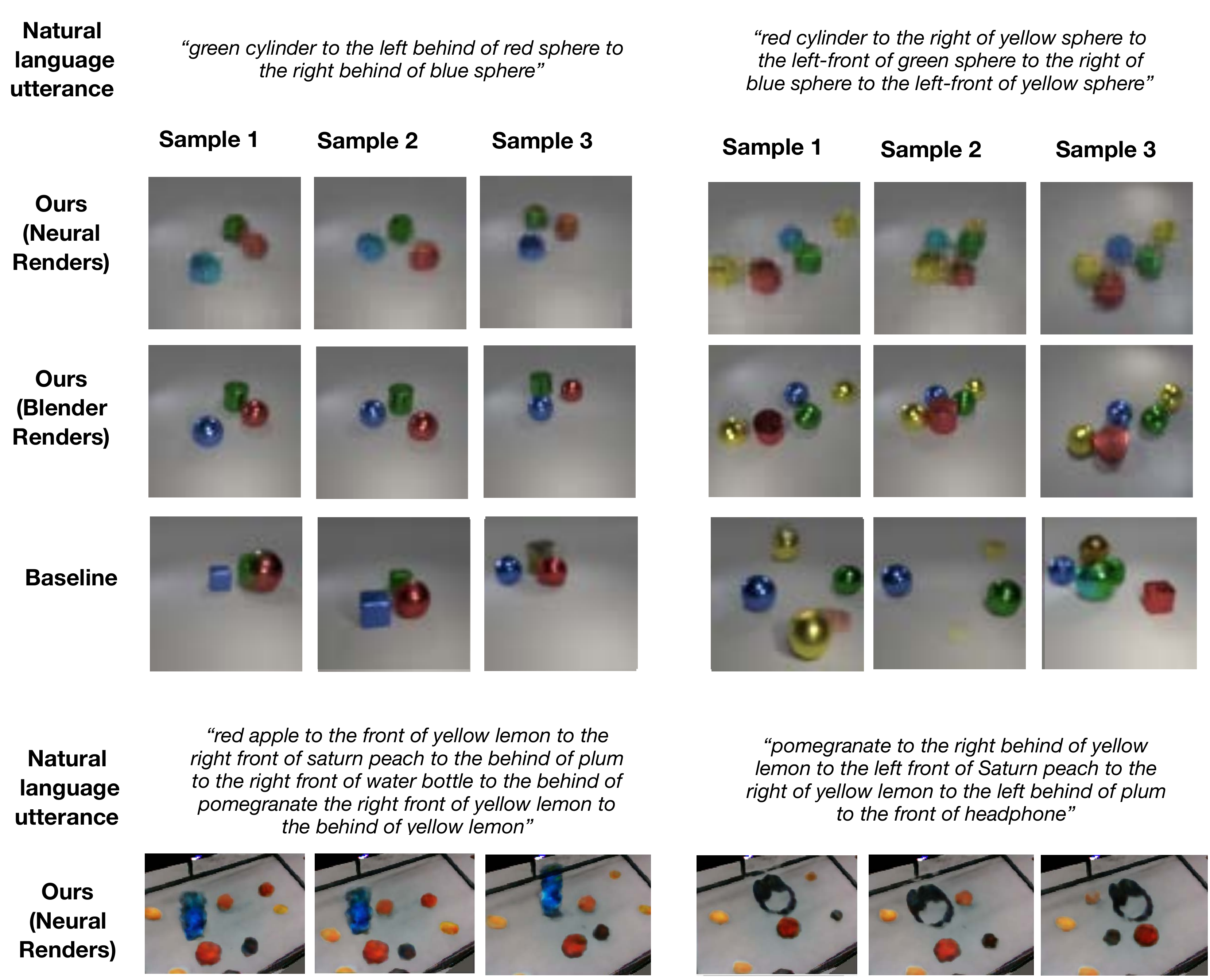}
    \caption{ 
    \textbf{Language to scene generation} (Rows 1,2,4) and \textbf{language to image generation} (Row 3) from our model and the model of Deng et al \cite{NIPS2018_7658}  for utterances longer than those encountered at training time on CLEVR and our real-world dataset. 
     Both our model and the baseline are stochastic, and we sample three generated scenes per utterance. 
    }
    \label{fig:generation}
\end{figure}

We test the proposed language grounding model in the following tasks: \textbf{(i)} Generating scenes based on language utterances
\textbf{(ii)} classifying utterances based on whether they describe possible or impossible scenes,  
\textbf{(iii)} detecting spatial referential expressions, and, \textbf{(iv)} following object placement instructions. 
We consider two datasets: (i) The CLEVR dataset of Johnson et al.~\cite{DBLP:journals/corr/JohnsonHMFZG16} that contains 3D scenes annotated with natural language descriptions, their dependency parse trees, and the object 3D bounding boxes. The dataset contains Blender generated 3D scenes with geometric objects. Each object can take a number of colors, materials, shapes and sizes. Each scene is accompanied with a description of the object spatial arrangements, as well as its parse tree. 
Each scene is rendered from 12 azimuths and 4 elevation angles, namely, 
$\{12^o,20^o,40^o,60^o\}$. 
We train GRNNs for view prediction using the RGB image views in the training sets. The annotated 3D bounding boxes are used to train our 3D  object detector.
We generate 800 scenes for training, and 400 for testing. The language is generated randomly with a \textbf{maximum of 2 objects} for the training scenes. 
(ii) A dataset we collected in the real world. We  built a camera dome comprised of 8 cameras placed in a hemisphere above a table surface. We move vegetables around and collect multiview images. We automatically annotate the scene with 3D object boxes by doing 3D pointcloud subtraction at training time, We use the obtained 3D boxes  to train our 3D object detector. At test time, we detect objects from a single view using our trained 3D detector.  We further provide category labels for the vegetable present in single-object scenes to facilitate the association of labels to object 3D bounding boxes. More elaborate multiple instance learning techniques could be used to handle the general case of weakly annotated multi-object scenes \cite{Mao2019NeuroSymbolic}.  We leave this for future work. We show extensive qualitative results on our real-world dataset as evidence that our model can effectively generalize to real-world data if allowed multiview embodied supervision and weak category object labels.



%

\subsection{Language conditioned scene generation} 
We show language-conditioned generated scenes for our model and the baseline model of Deng et al.~\cite{NIPS2018_7658} in Figure \ref{fig:generation} for utterances longer than those encountered at training time.  The model of Deng et al.~\cite{NIPS2018_7658} generates a 2D RGB image directly (without an intermediate 3D representation) conditioned on a  language utterance and its dependency tree. For each object mentioned in the utterance, the model of Deng et al.~\cite{NIPS2018_7658} predicts the absolute 2D location, 2D box size and a 2D appearance feature map for the object, and then it warps and places the 2D appearance feature map on a canvas according to the predicted location and object sizes. The canvas with 2D features is neurally decoded into an RGB image.
We visualize our own model's predictions in two ways: i) \textbf{neural renders}   are obtained by feeding the generated 3D assembled canvas to the 3D-to-2D neural projection  module of GRNNs, ii) \textbf{Blender renders} are renderings of Blender  scenes that contain 3D meshes   
selected by nearest neighbor to  the language generated  object 3D feature tensors, and arranged based on the 
predicted 3D spatial offsets.

Our model re-samples an object location when it detects that the newly added object penetrates the existing objects, with a 3D intersection-over-union (IOU) score higher than a cross-validated threshold of 0.1. 
The model of Deng et al.~\cite{NIPS2018_7658} is trained to handle occluded objects. Notice in Figure \ref{fig:generation} that it generates weird configurations as the number of objects increase. We tried imposing constraints of object placement using 2D IoU threshold in our baseline, but ran into the problem that we could not find plausible configurations for strict IoU thresholds, and we would obtain non-sensical configurations for low IoU thresholds,  we include the results in the supplementary file. 
Note that 2D IoU cannot discriminate between physically plausible object occlusions and physically implausible object intersection. Reasoning about 3D object non intersection is indeed much easier in 3D space. 


Sections 2 and 3 of the supplementary file include more scene generation examples, where predictions of our model are decoded from multiple camera viewpoints, more comparisons against the baseline, and more details on the Blender rendering visualization. Please note that \textbf{image generation is not the end-task for this work; instead, it is a task to help learn the mapping from language to the 3D space-aware feature space.}  
We opt for a model that has reasoning capabilities over the generated entities, as opposed to  generating pixel-accurate images that we cannot reason on.

\subsection{Affordability inference of natural language utterances} \label{sec:afford}
We  test our model and baselines  in their ability to classify language utterances as describing sensical or non-sensical object configurations. We created a test set of 92 NL utterances, 46 of which are affordable, i.e., describe a realizable object arrangement, e.g., \textit{``a red cube is in front of a blue cylinder and in front of a red sphere, the blue cylinder is in front of the red sphere.''}, and 46 are unaffordable, i.e., describe a non-realistic object arrangement, e.g.,  \textit{``a red cube is behind a cyan sphere and in front of a red cylinder, the cyan sphere is left behind the red cylinder''}.  In each utterance, an object is mentioned multiple times. The utterance is unaffordable when these mentions are contradictory. 
Answering correctly requires spatial reasoning over possible object configurations. 
\textbf{Both our model and the baselines have been trained only on plausible utterances and scenes. We use our  dataset only for evaluation. } This setup is similar to violation of expectation \cite{Riochet2019IntPhysAB}: the model detects violations while it has only been trained on plausible versions of the world.

Our model infers affordability of a language utterance by generating the 3D feature map of the described scene, as detailed in Section \ref{sec:simsem}.  
When an object is mentioned multiple times in an utterance, our model uses the first mention to add it in the 3D feature canvas, and uses the pairwise object spatial classifier $\DS$ of Section \ref{sec:refdet} to infer if the predicted configuration also satisfies the later constraints. If not, our model re-samples object arrangements until a configuration is found or a maximum number of samples is reached. 

We compare  our model against a baseline based on the model of Deng et al.~\cite{NIPS2018_7658}.  Similar to our model, when an object is mentioned multiple times, we use the first mention to add it in the 2D image canvas, and use  pairwise object spatial classifiers we train over 2D bounding box spatial information---as opposed to 3D---to infer if the predicted configuration also satisfies the later constraints. Note that there are no previous works that attempt this language  affordability inference task, and our baseline essentially performs similar operations as our model but in a 2D space.

We consider a sentence to be affordable if the spatial classifier predicts a score above 0.5 for the later constraint. \textbf{Our model achieved an affordability classification accuracy of $95\%$ while the baseline achieved $79\%$}. 
This suggests that reasoning in 3D as opposed to 2D makes it easier to determine the affordability of object configurations. 

\subsection{Detecting spatial referential  expressions}
 To evaluate our model's ability to detect spatial referential expressions, we use the same dataset and train/test split of scenes as in the previous section. 
 For each annotated scene, we consider the first mentioned object as the one being referred to, that needs to be detected.
In this task, we compare our model with a variant of the modular 2D referential object detector of Hu et al.~\cite{modularreferential} that also takes as input the dependency parse tree of the expression.  We train the object appearance detector for the baseline the same way as we train our model using positive and negative examples, but the inner product is on 2D feature space as opposed to 3D. We also train a pairwise spatial expression classifier to map width, height and x,y coordinates of the two 2D bounding boxes and the one-hot encoding of the spatial expression, e.g., \textit{``in front of"}, to a score reflecting whether the two boxes respect the corresponding arrangement. Note that our pairwise spatial expression classifier uses 3D box information instead, which helps it to generalize across camera placements.

\begin{small}
\begin{table}
  \vspace{-1em} 
\centering

\begin{tabularx}
{\textwidth}{|p{0.10\textwidth}|p{0.19\textwidth}|p{0.17\textwidth}|p{0.15\textwidth}|p{0.132\textwidth}|}
\hline {\footnotesize mAP}  & {\footnotesize ours RGB-D}  &  {\footnotesize \cite{NIPS2015_5638} RGB-D} & {\footnotesize ours RGB} & {\footnotesize \cite{NIPS2015_5638} RGB} \\ \hline
2D & \textbf{0.993}     &  0.903 & 0.990     &  0.925  \\ \hline  3D    &  \textbf{0.973}    & -  & 0.969   & - \\ \hline 
\end{tabularx}
\caption{\textbf{Mean average precision for category agnostic region proposals on Clevr dataset.} Our 3D RPN outperforms the 2D state-of-the-art RPN of Faster R-CNN \cite{NIPS2015_5638}.} \label{tab:rpn}
\end{table}
\end{small}

Our referential detectors are upper bounded by the performance of the Region Proposal Networks (RPNs) in 3D for our model and in 2D for the baseline, since we use language-generated object feature tensors  to compare with object features extracted from 2D and 3D bounding box proposals. We evaluate RPN performance in Table \ref{tab:rpn}.    An object is  successfully detected when the predicted box has an  intersection over union (IoU) of at least 0.5 with the groundtruth bounding box. For our model, we project the detected 3D boxes to 2D and compute 2D mean average precision (mAP). Both our model and the baseline use a single RGB image as input as well as a corresponding depth map, which our model uses during the 2D-to-3D unprojection operation and the 2D RPN concatenates with the RGB input image. Our 3D RPN that takes the GRNN map $\M$ as input better delineates the objects under heavy occlusions than the 2D RPN.  

\begin{figure}[t!]
    \centering
    \includegraphics[width=1.0\textwidth]{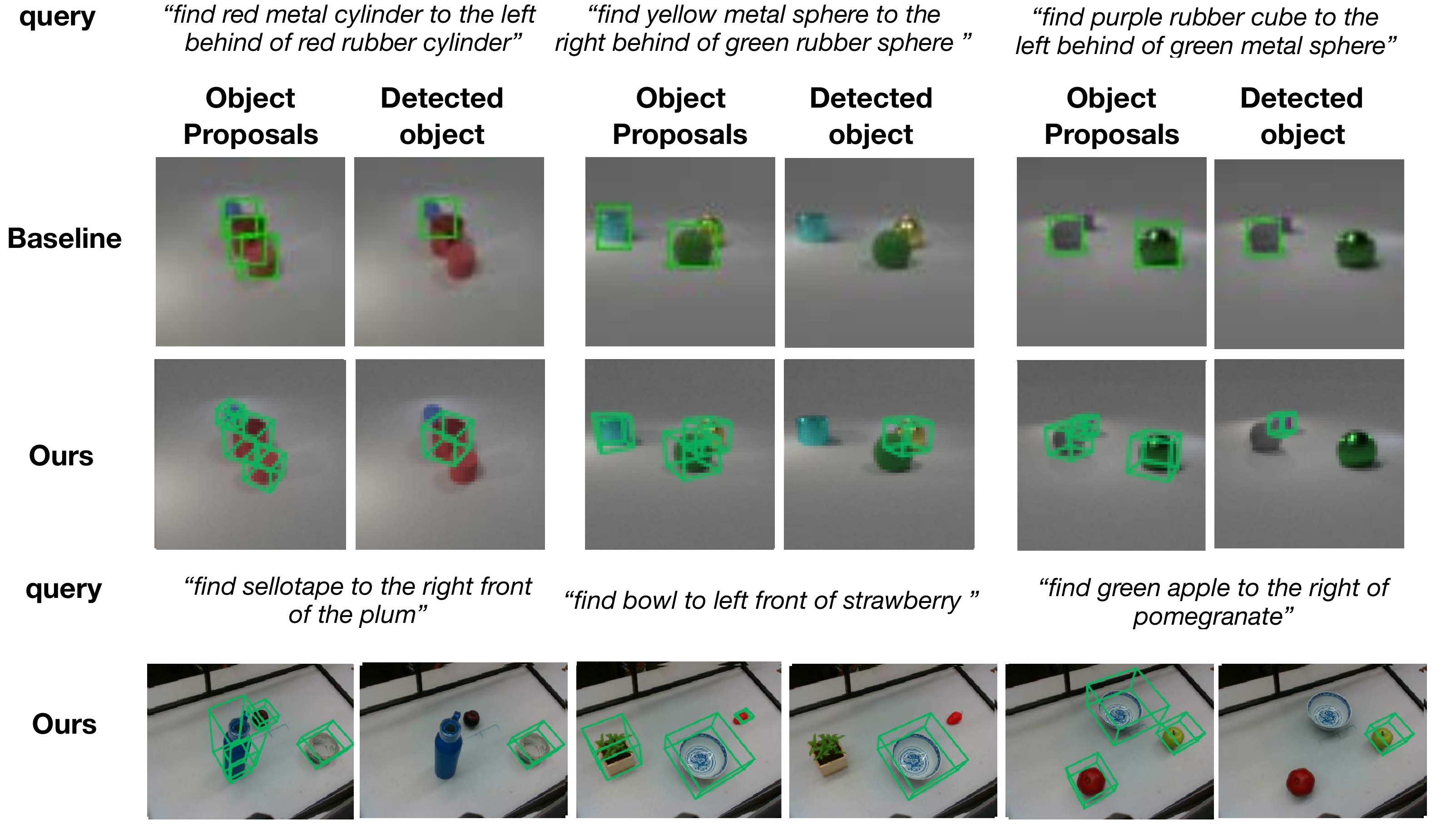}
    \caption{ 
    \textbf{Detecting  referential spatial expressions.} On Clevr and our real-world dataset, we show given a scene and a referential expression, our model localizes the object being referred to in 3D, while our  baseline in 2D.
    }
    \label{fig:referentials}
\end{figure}

We show quantitative results for referential expression detection in Table \ref{tab:referentials} with groundtruth as well as RPN predicted boxes, and qualitative results in Figure \ref{fig:referentials}. In the \textit{``in-domain view"} scenario, we test on camera viewpoints that have been seen at training time, in the \textit{``out-of-domain view"} scenario, we test on \textbf{novel camera viewpoints.} 
An object is detected successfully when the corresponding  detected bounding box has an IoU  of 0.5 with the groundtruth box (in 3D for our model and in 2D for the baseline). Our model greatly outperforms the baseline for two reasons: a) it better detects objects in the scene despite heavy occlusions, and, b) even with groundtruth boxes, our model generalizes  better across camera viewpoints and object arrangements because the 3D representations of our model do not suffer from projection artifacts.

\begin{table}[h!]
\centering
\begin{tabularx}{\textwidth}{|p{0.28\textwidth}|p{0.06\textwidth}|p{0.06\textwidth}|p{0.17\textwidth}|p{0.17\textwidth}|}
\hline
  & Ours  &  \cite{modularreferential}  & Ours - GT 3D boxes  & \cite{modularreferential} - GT 2D boxes \\ \hline
\textit{in-domain view}   & 0.87  & 0.70 & \textbf{0.91}     & 0.79 \\ \hline
\textit{out-of-domain view}   & 0.79 & 0.25 & \textbf{0.88}     & 0.64 \\ \hline
\end{tabularx}
\caption{
\textbf{F1-Score for detecting  referential expressions.} 
Our model greatly outperforms the baseline  with both groundtruth and  predicted region proposals, especially for novel camera views on the CLEVR dataset. 
}
\label{tab:referentials}
\end{table}

\subsection{Manipulation instruction following} \label{sec:instrfolexp}


We use the PyBullet Physics simulator \cite{bullet} with simulated KUKA robot arm as our robotic platform. 
We  use  a \textit{cube} and a \textit{bowl}, using the same starting configuration for each scene, where the cube is held right above the bowl. 
We fix the end-effector to always point downwards.
 
 We compare our model against 
the 2D generative baseline of \cite{NIPS2018_7658} that generates object locations in 2D, and thus supply costs of the form:  $\mathcal{C}^{2D}(\mathbf{x}_t)=\|\mathbf{x}^{2D}_t-\mathbf{x}^{2D}_{goal}\|^2_2$. We show in Table \ref{tab:placement2} success rates for different spatial expressions, where we define success as placing the object in the set of  locations implied by the instruction. 
 Goal locations provided in 2D do much worse in guiding policy search than target object locations in 3D supplied by our model. This is because 2D distances suffer from foreshortening and reflect planning distance poorly. This is not surprising: in fact, the robotics control literature almost always considers desired locations of objects to be achieved to be in 3D \cite{DBLP:journals/corr/KumarGTL16,Levine:2016:ETD:2946645.2946684}. In our work, we link language instructions with such 3D inference using inverse graphics computer vision architectures for 2D to 3D lifting in a learnable 3D feature space. 
Videos of the learnt language-conditioned placement policies can be found here: \url{https://mihirp1998.github.io/project_pages/emblang/}
\begin{table}[h!]
\begin{tabularx}{\textwidth}{|p{0.145\textwidth}|p{0.035\textwidth}|p{0.09\textwidth}|p{0.065\textwidth}|p{0.053\textwidth}|p{0.092\textwidth}|p{0.07\textwidth}|p{0.03\textwidth}|}
\hline
Language Exp. &  left & left-behind & left-front &right &right-behind & right-front & in \\ \hline
{Baseline} &  4/5 &1/5 &3/5 & 0/5 & 2/5 & 0/5 & 1/5 \\ \hline
{Ours} &  \textbf{5/5} &\textbf{3/5} &\textbf{5/5} & \textbf{5/5} & \textbf{5/5} & \textbf{3/5} & \textbf{5/5} \\ \hline
\end{tabularx}
\caption{
\textbf{Success rates for executing instructions regarding object placement.} Policies learnt 
using costs over 3D configurations much outperform those learnt with costs over 2D configurations. }
\label{tab:placement2}
\end{table}

\vspace{-0.1in}
\vspace{-0.1in}

\section{Discussion - Future Work} \label{sec:lim}
We proposed models that 
associate language utterances  with compositional 3D feature representations of the objects and scenes the utterances describe, and exploit the rich constrains of the 3D space for spatial reasoning. 
We showed our model can effectively imagine object spatial configurations conditioned on language utterances,  can reason about affordability of spatial arrangements, detect objects in them, and train policies for following object placement instructions. We further showed our model generalizes to real world data without real world  examples of scenes annotated with spatial descriptions, rather, only single category labels. 
The language utterances we use are  programmatically generated \cite{DBLP:journals/corr/JohnsonHMFZG16}. One way to extend our framework to handle truly natural language is by  paraphrasing such programmatically generated utterances \cite{conf/acl/BerantL14} to create paired examples of natural language utterances and parse trees, then  train a dependency parser \cite{DBLP:journals/corr/WeissACP15} to generate dependency parse trees as input for our model using natural language as input. 
Going beyond basic spatial arrangements would require learning dynamics,  physics and mechanics of the grounding 3D feature space. 
These are clear avenues for future work.


\section{Acknowledgements}
We would like to thank Shreshta Shetty and Gaurav Pathak for helping set up the table dome and  Xian Zhou for help with the real robot placement experiments. This work was partially funded by Sony AI and AiDTR grant. Hsiao-Yu Tung is funded by Yahoo InMind Fellowship and Siemens FutureMaker Fellowship.

{\small
\bibliographystyle{ieee_fullname}
\bibliography{8_refs}
}

\clearpage
{

\section{Model details: Language-conditioned 3D visual imagination} 
We train our stochastic generative networks using conditional variational autoencoders. 
For the \textit{what} generative module, our inference network conditions on the word embeddings of the adjectives and the noun in the noun phrase,
as well as the 3D feature tensor obtained
by cropping the 3D feature map $\M=\GRNN(I)$ using the  ground-truth 3D  bounding box of the object the noun phrase concerns.  
For the \textit{where} generative module, the corresponding inference network conditions on one-hot encoding of the spatial expression, as well as the 3D relative spatial offset, available from 3D object box annotations. Inference networks are used only at training time. 
Our {\it what} and {\it where} decoders take the posterior noise and predict 3D object appearance feature tensors, and cross-object 3D spatial offsets, respectively, for each object. We add predicted object feature tensors at predicted 3D locations in a 3D feature canvas. 
Our reconstruction losses ask the language-generated  and image-inferred 3D feature maps from GRNNs to be close in feature distance, both in 3D and after 2D neural projection using the GRNN 3D-to-2D neural decoder, and the predicted cross-object 3D relative spatial offsets to be close to the ground-truth cross-object 3D  relative offsets.

\section{Experimental details: Language conditioned scene generation}
We visualize our model's predictions in two ways: i) \textbf{neurally rendered}   are obtained by feeding the generated 3D assembled canvas to the 3D-to-2D neural projection  module of GRNNs, ii) \textbf{Blender rendered} are renderings of Blender  scenes that contain  object 3D meshes   
selected by small feature distance to  the language generated  object 3D feature tensors, and arranged based on the 
predicted 3D spatial offsets.

We consider a database of 300  object 3D meshes to choose from. 
To get the object feature tensor for a candidate 3D object model, we render multi-view RGB-D data of this object in Blender, and input them to the GRNN to obtain the corresponding feature map,
which we  crop using the groundtruth bounding box.
Blender renders better convey object appearance because the neurally rendered images are blurry. Despite pixel images being blurry, our model retrieves correct object meshes that match the  language descriptions.

\section{Additional experiments}

\paragraph{Scene generation conditioned on natural language} In Figure \ref{fig:samplegen1}, we compare our model with the model of Deng et al \cite{NIPS2018_7658} on language to scene generation with utterances longer than those used during training time.
We show both neural and Blender rendering of scenes predicted from our model. We remind the reader that a Blender rendering is computed by using the cross-object relative 3D offsets predicted by our model, and using the generated object 3D feature tensors to retrieve the closest matching meshes from a training set. Our training set is comprised of 100 objects with known 3D bounding boxes, and for each we compute a 3D feature tensor by using the 2D-to-3D unprojection module described above, and cropping the corresponding sub-tensor based on the 3D bounding box coordinates of the object. Despite our neural rendering being blurry, we show the features of our generative networks achieve correct nearest neighbor retrieval. The generation results show our model can generalize to utterances that are much longer than those in the training data. In Figure \ref{fig:samplegen2}, we show rendering results from our model on our real world dataset.

One key feature of our model is that it generates a scene as opposed to an independent static image. In Figure \ref{fig:generation_angle}, we show rendering images from the 3D feature tensor across different viewpoints. The rendering images are consistent across viewpoints. For a 2D baseline \cite{NIPS2018_7658}, it is unclear how we can obtain a set of images that not only match with input sentence but also are consistent with each others.

We show in Figures~\ref{fig:scenegen1}-\ref{fig:scenegen2} more neural and Blender rendering of scenes predicted from our model, conditioning on parse trees of natural language utterances.  In \ref{fig:scenegen3}, we show rendering results learned from our real world dataset.

\paragraph{Scene generation conditional on natural language and  visual context}
In Figures~\ref{fig:conditional1}-\ref{fig:conditional3} we show examples of  scene generation from our model when conditioned on both natural language and the visual context of the agent.  In this case, some objects mentioned in the natural language utterance are present in the agent's environment, and some are not.  Our model uses a 3D object detector to localize objects in the scene, and the learnt 2D-to-3D unprojection neural module to compute a 3D  feature tensor for each, by cropping the scene tensor around each object. Then, it compares the object tensors generated from natural language to those generated from the image, and if a feature distance is below a threshold, it grounds the object reference in the parse tree of the utterance to object present in the environment of the agent. 
If such binding occurs, as is the case  for the ``green cube" in the top left example, then our model uses the image-generated tensors of the binded objects, instead of the natural language generated ones, to complete the imagination. In this way, our model grounds natural language to both perception and imagination.

\paragraph{Affordability inference based on 3D non-intersection}
Objects do not intersect in 3D. Our model has a 3D feature generation space and can detect when this basic principle is violated. The baseline model of \cite{NIPS2018_7658} directly generates 2D images described in the utterances (conditioned on their parse tree) without an intermediate 3D feature space. Thus, it performs such affordability checks in 2D. However, in 2D, objects frequently occlude one another, while they still correspond to an affordable scene. In Figure \ref{fig:affordability}, we show intersection over union scores computed in 3D by our model and in 2D by the baseline. While for our model such scores correlate with affordabilty of the scene (e.g., the scenes in 1st, third, and forth columns in the first row are clearly non-affordable as objects inter-penetrate) the same score from the baseline is not an indicator of affordability, e.g., the last column in the last row of the figure can in fact be a perfectly valid scene, despite the large IoU score.

\paragraph{Language-guided placement policy learning}
In Figure \ref{fig:placement1}, we show the initial and final configurations of the learned policy using different referential expression. The robot can successfully place the object to the target location given the referential expression. We also show in the supplementary a video of a real robot executing the task.


\begin{figure*}[h]
    \centering
     \includegraphics[width=1.0\textwidth]{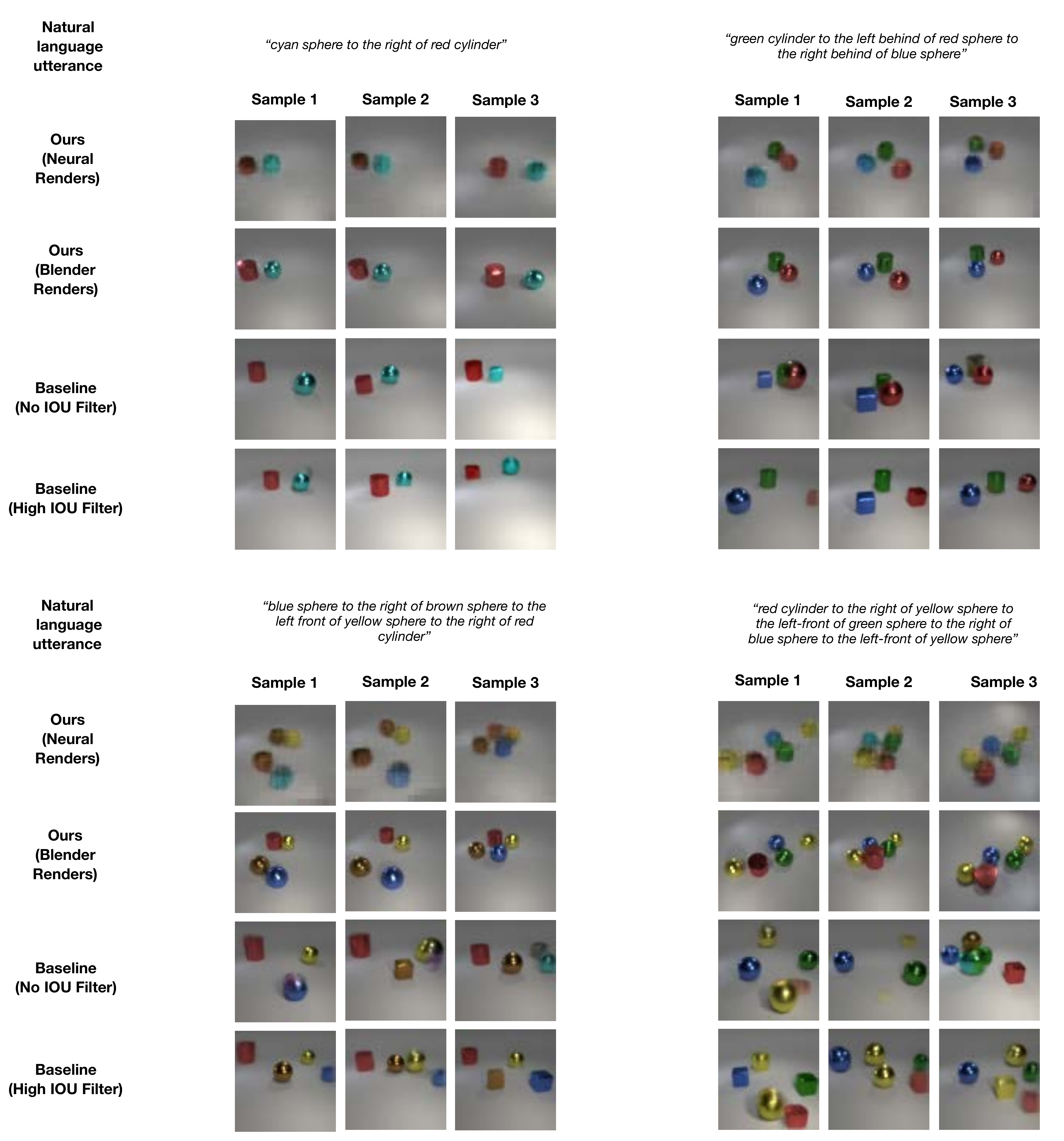}
\caption{
    \textbf{Language to scene generation} from our model (Row1, Row2) and the model of Deng et al \cite{NIPS2018_7658} (Row 3, 4) for utterances longer than those encountered at training time.
     Both our model and the baseline are stochastic, and we sample three generated scenes/images per utterance. (Row 1 and Row 2) shows neural and Blender rendering results from our model. For the Blender rendering, we retrieve the closet 3D object mesh using the features of our generative networks, and place the retrieved objects in the corresponding locations in Blender to render an image.
     (Row 3) shows  image generation results with no IoU constraint during sampling for the baseline. This means objects might go out of the field of view. (Row 4) shows
     result with high IoU constraint. 
    }
    \label{fig:samplegen1}
\end{figure*}

\begin{figure*}[h]
    \centering
     \includegraphics[width=1.0\textwidth]{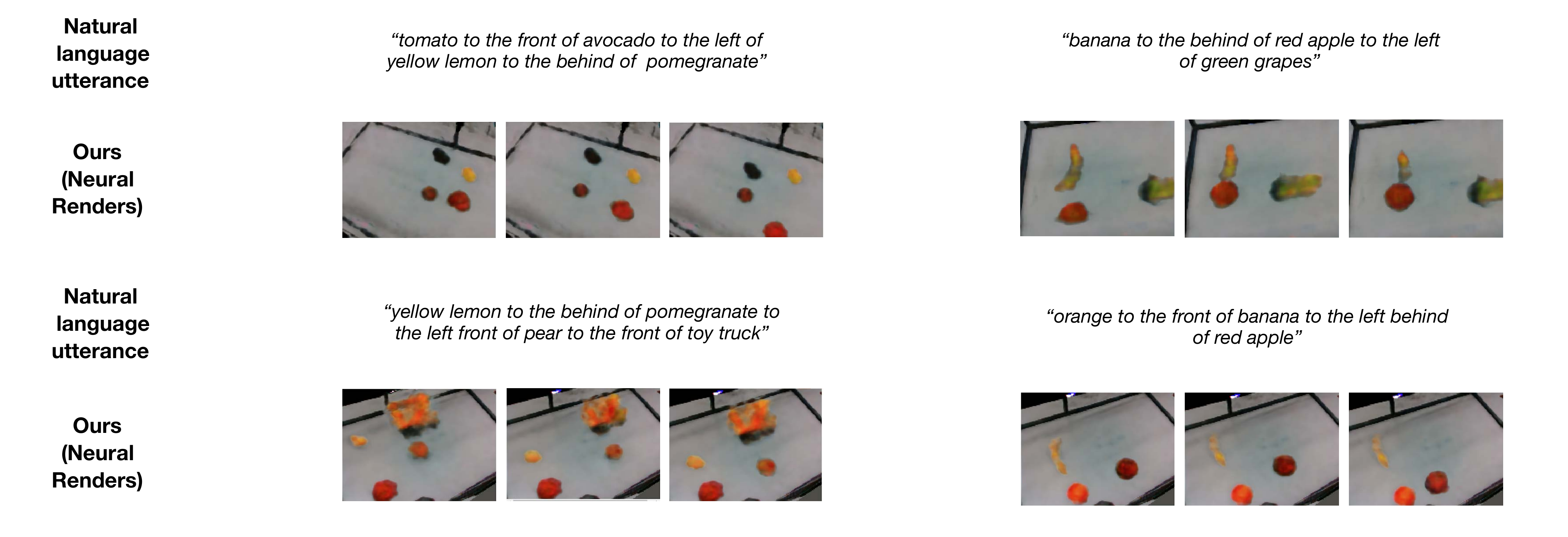}
\caption{
     \textbf{Language to image generation} on our real world data. We sample three different scenes for each natural language utterances.
    }
    \label{fig:samplegen2}
\end{figure*}

\begin{figure*}[t!]
    \centering
    \includegraphics[width=1.0\textwidth]{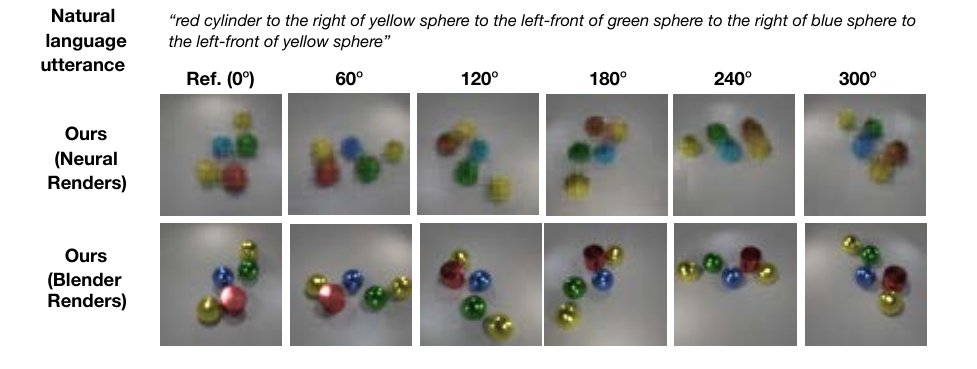}
    \caption{ 
    \textbf{Consistent scene generation} . We render the generated 3D feature canvas from various viewpoints in the first row using the neural GRNN decoder, and compare against the different viewpoint projected Blender rendered scenes. Indeed, our model correctly predicts occlusions and visibilities of objects from various viewpoints, and can generalize across different number of objects. 2D baselines do not have such imagination capability. 
    }
    \label{fig:generation_angle}
\end{figure*}

\begin{figure*}[h]
    \centering
     \includegraphics[width=\textwidth]{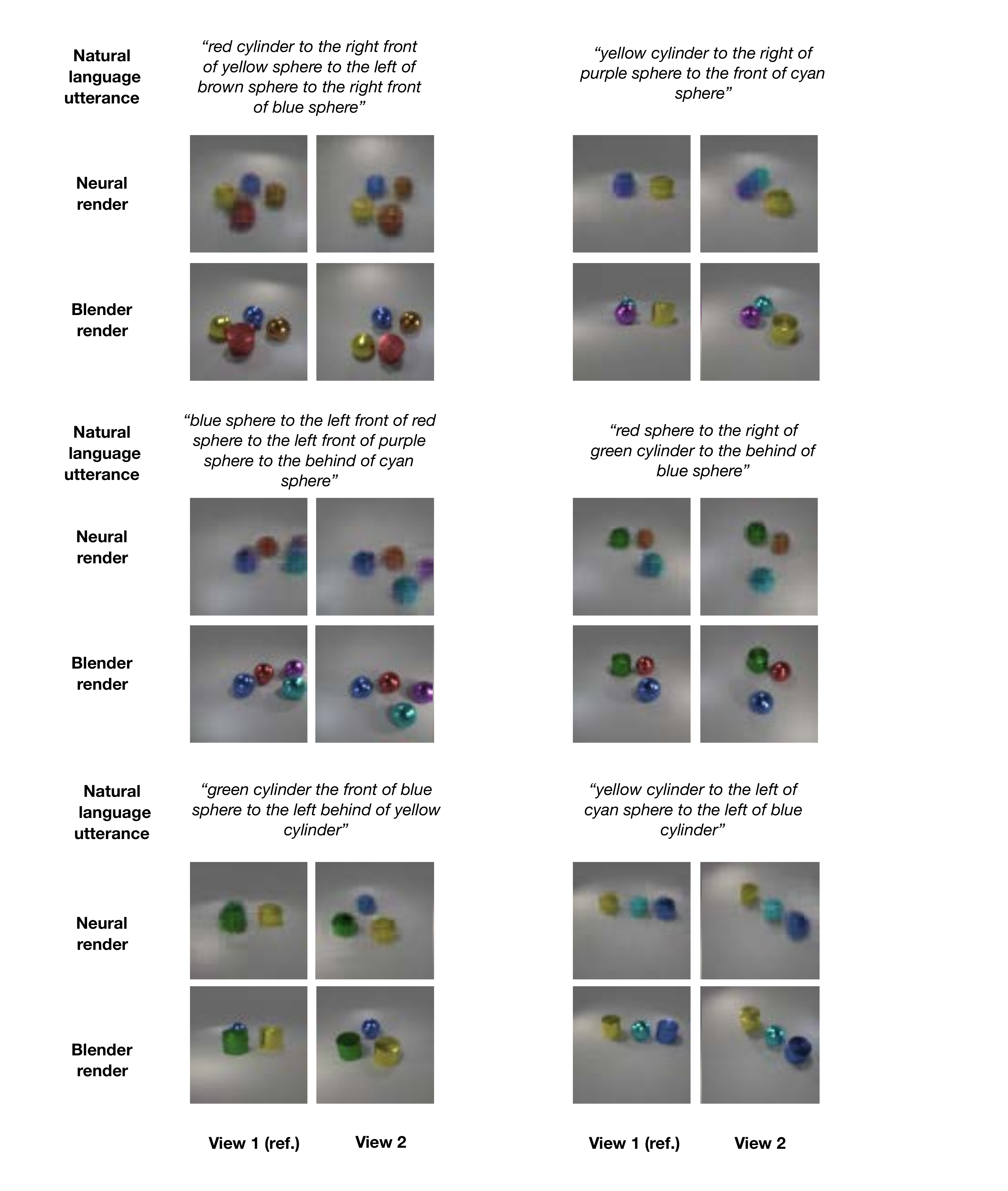}
    \caption{\textbf{ Natural language conditioned neural and blender scene renderings generated by the proposed model.}  We visualize each scene from two nearby views, a unique ability of our model, due to its 3-dimensional generation space.
    }
    \label{fig:scenegen1}
\end{figure*}
\begin{figure*}[h]
    \centering
    \includegraphics[width=\textwidth]{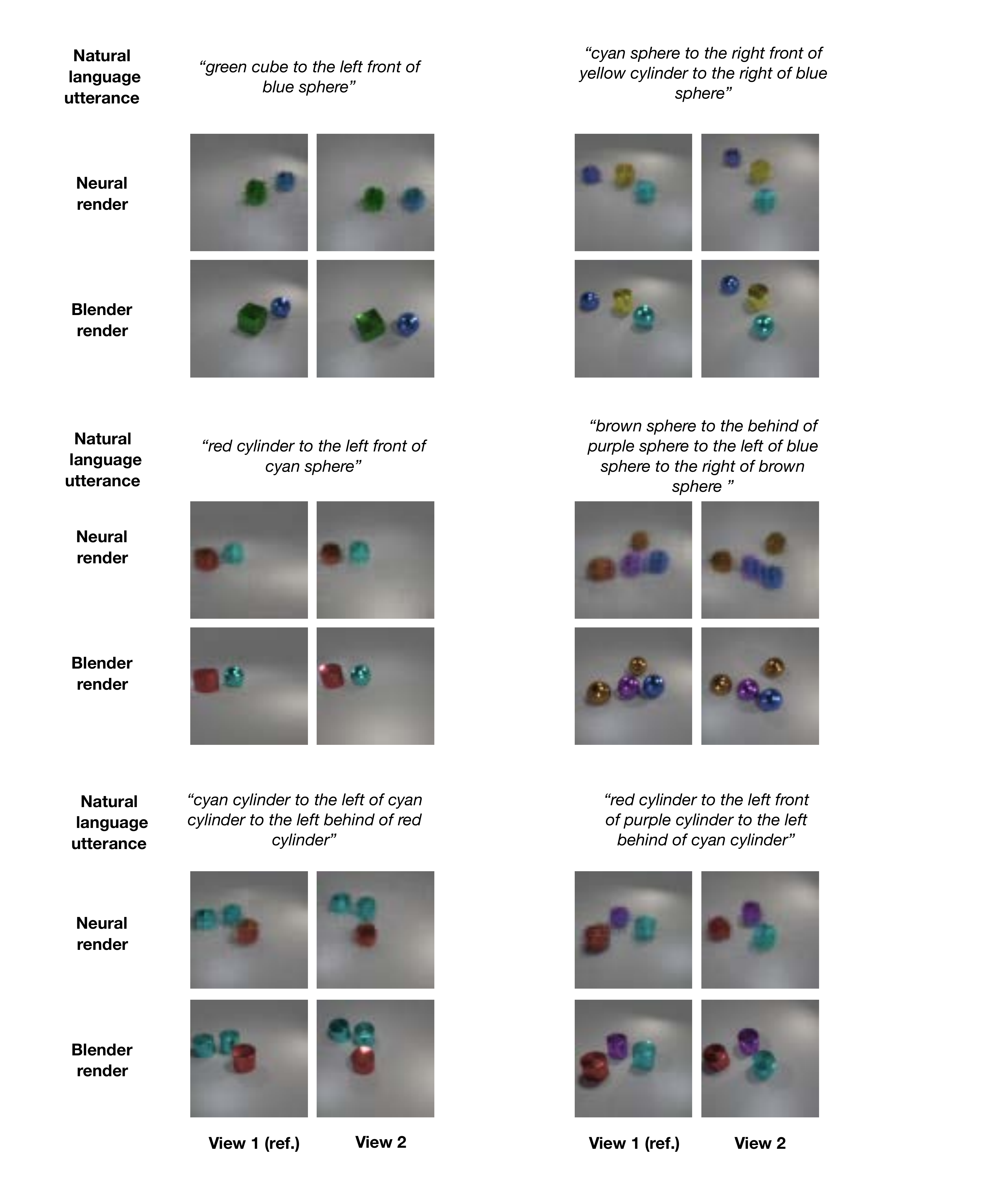}
    \caption{ \textbf{(Additional) Natural language conditioned neural and blender scene renderings generated by the proposed model.} 
    }
    \label{fig:scenegen2}
\end{figure*}
\begin{figure*}[h]
    \centering
    \includegraphics[width=0.85\textwidth]{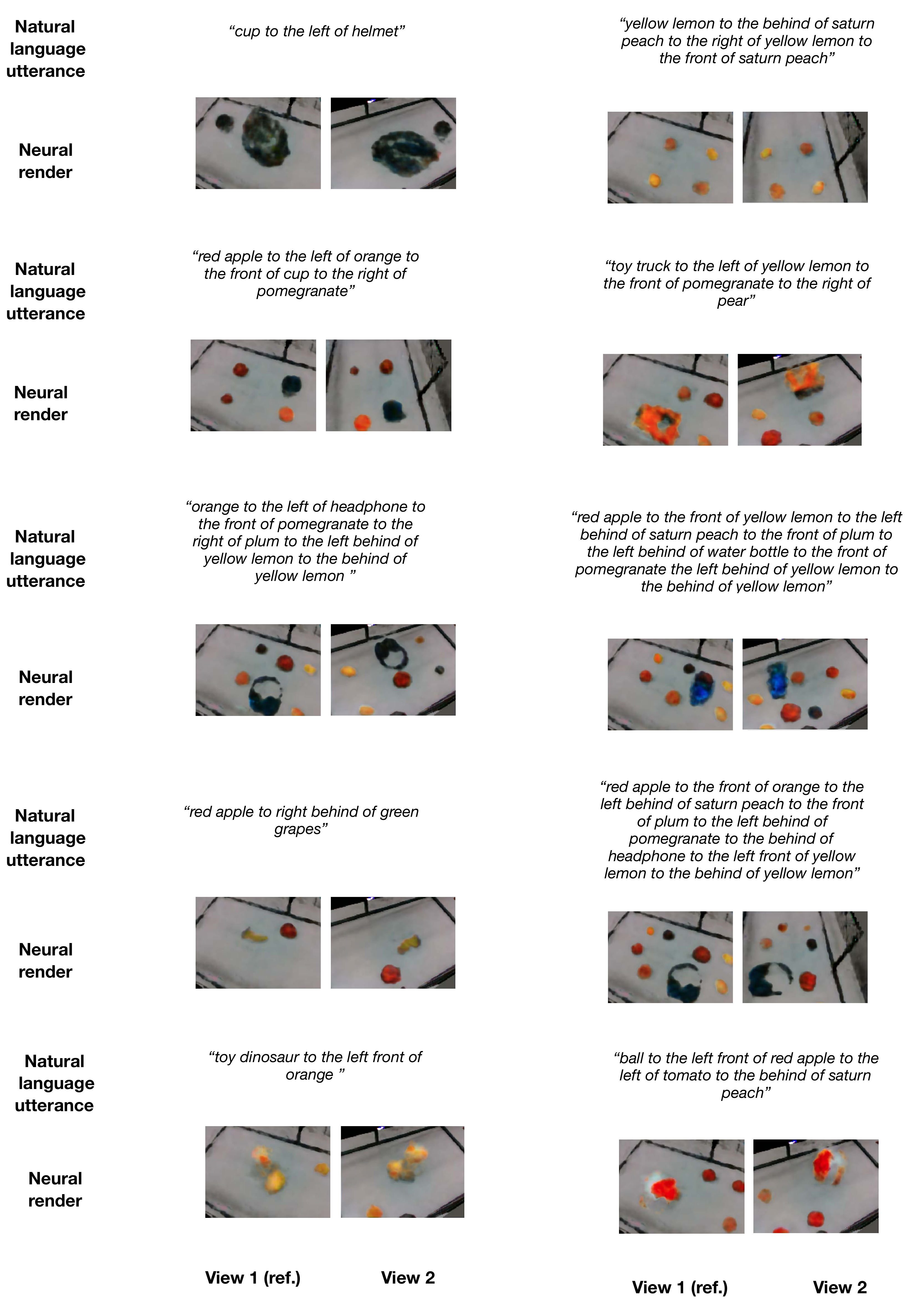}
    \caption{ \textbf{(Additional) Natural language conditioned neural scene renderings generated by the proposed model over our real world dataset.} 
    }
    \label{fig:scenegen3}
\end{figure*}

\begin{figure*}[h]
    \centering
    \includegraphics[width=\textwidth]{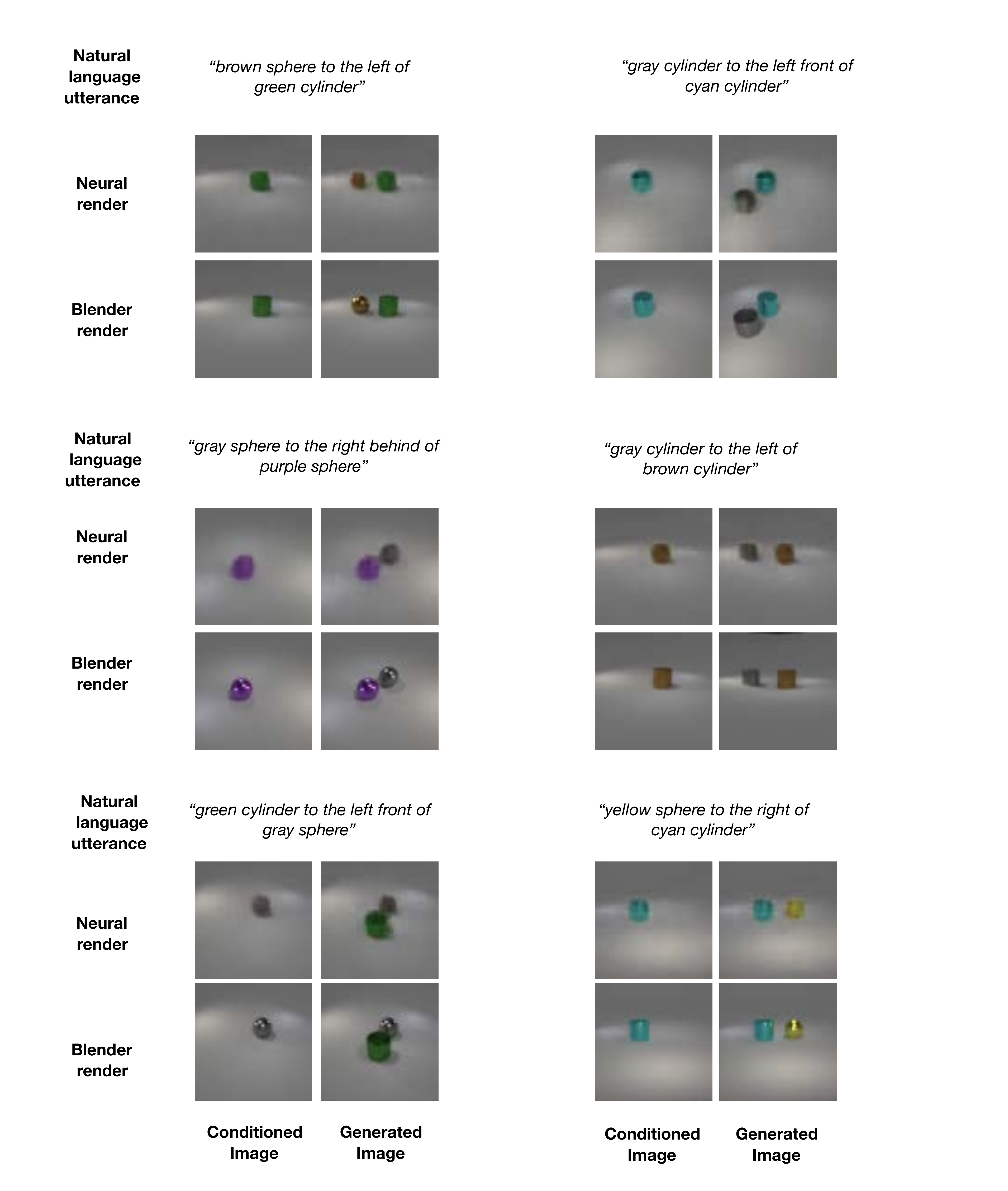}
    \caption{ \textbf{Neural and blender scene renderings generated by the proposed model, conditioned on natural language \textit{and} the visual scene.}  Our model uses a 3D object detector to localize objects in the scene, and the learnt 2D-to-3D unprojection neural module to compute a 3D  feature tensor for each, by cropping accordingly the scene tensor. Then, it compares the natural language conditioned generated object tensors to those obtained from the image, and  grounds  objects references in the parse tree of the utterance to objects presents in the environment of the agent, if the feature distance is below a threshold. If such binding occurs, as is the case  for the ``green cube" in top left, then, our model used the image-generated tensors of the binded objects, instead of the natural language generated ones, to complete the imagination. In this way, our model grounds natural language to both perception and imagination. }
    \label{fig:conditional1}
\end{figure*}

\begin{figure*}[h]
    \centering
    \includegraphics[width=\textwidth]{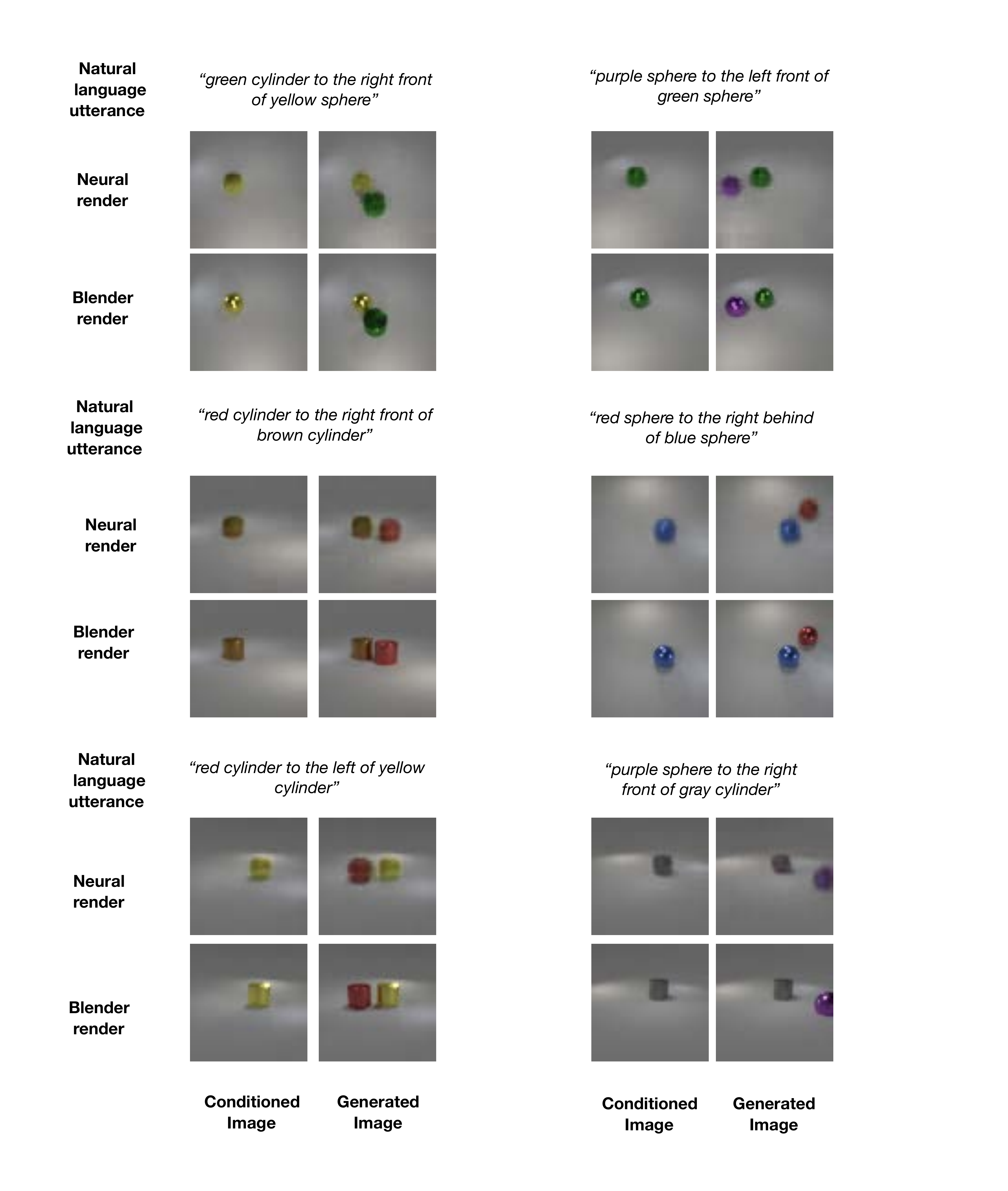}
    \caption{ \textbf{(Additional) {Neural and blender scene renderings generated by the proposed model, conditioned on natural language \textit{and} the visual scene.}} 
    }
    \label{fig:conditional2}
\end{figure*}
\begin{figure*}[h]
    \centering
    \includegraphics[width=0.8\textwidth]{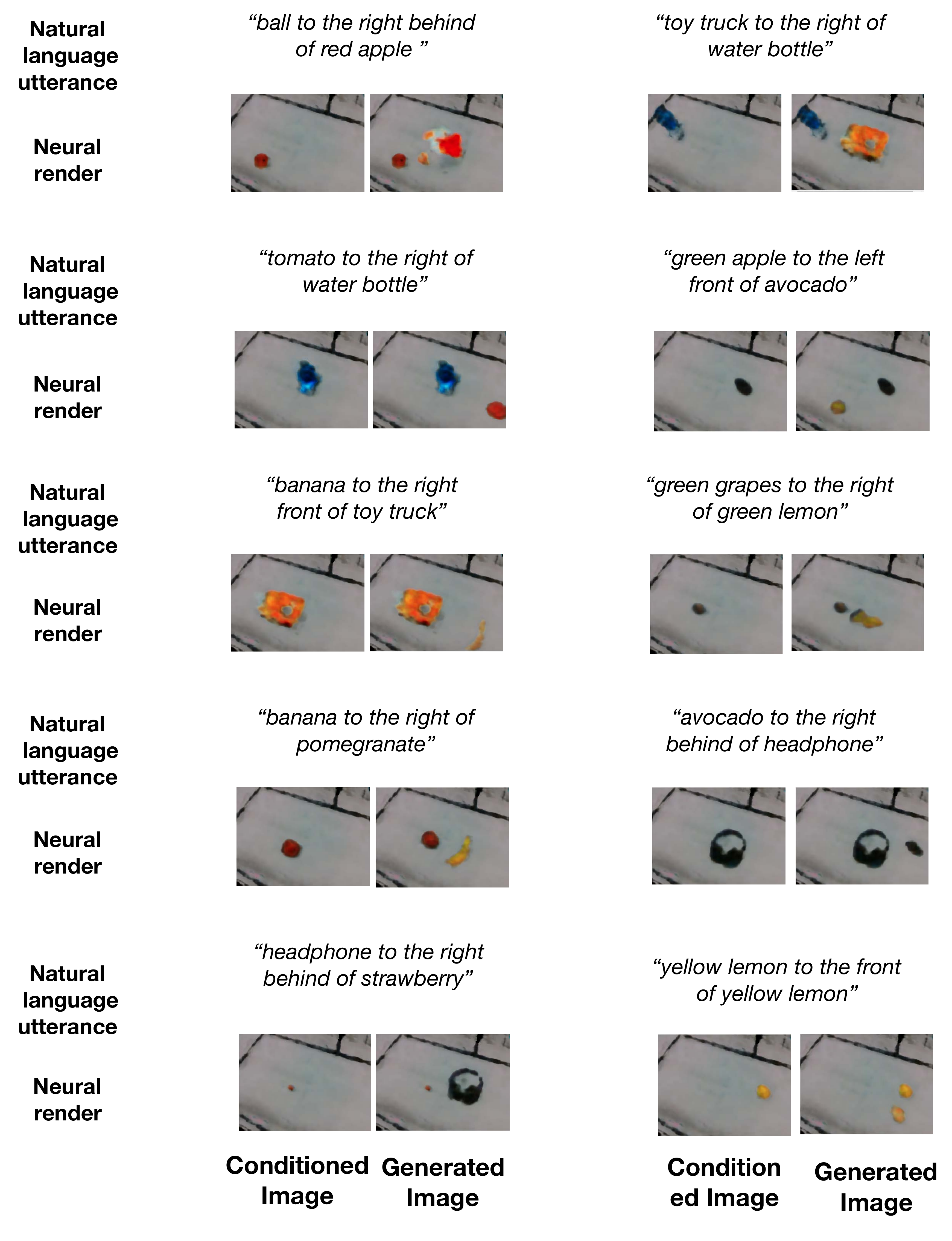}
    \caption{ \textbf{(Additional) {Neural scene renderings generated by the proposed model, conditioned on natural language \textit{and} the visual scene from our real world dataset.}} 
    }
    \label{fig:conditional3}
\end{figure*}
\begin{figure*}[t!]
    \centering
    \includegraphics[width=\textwidth]{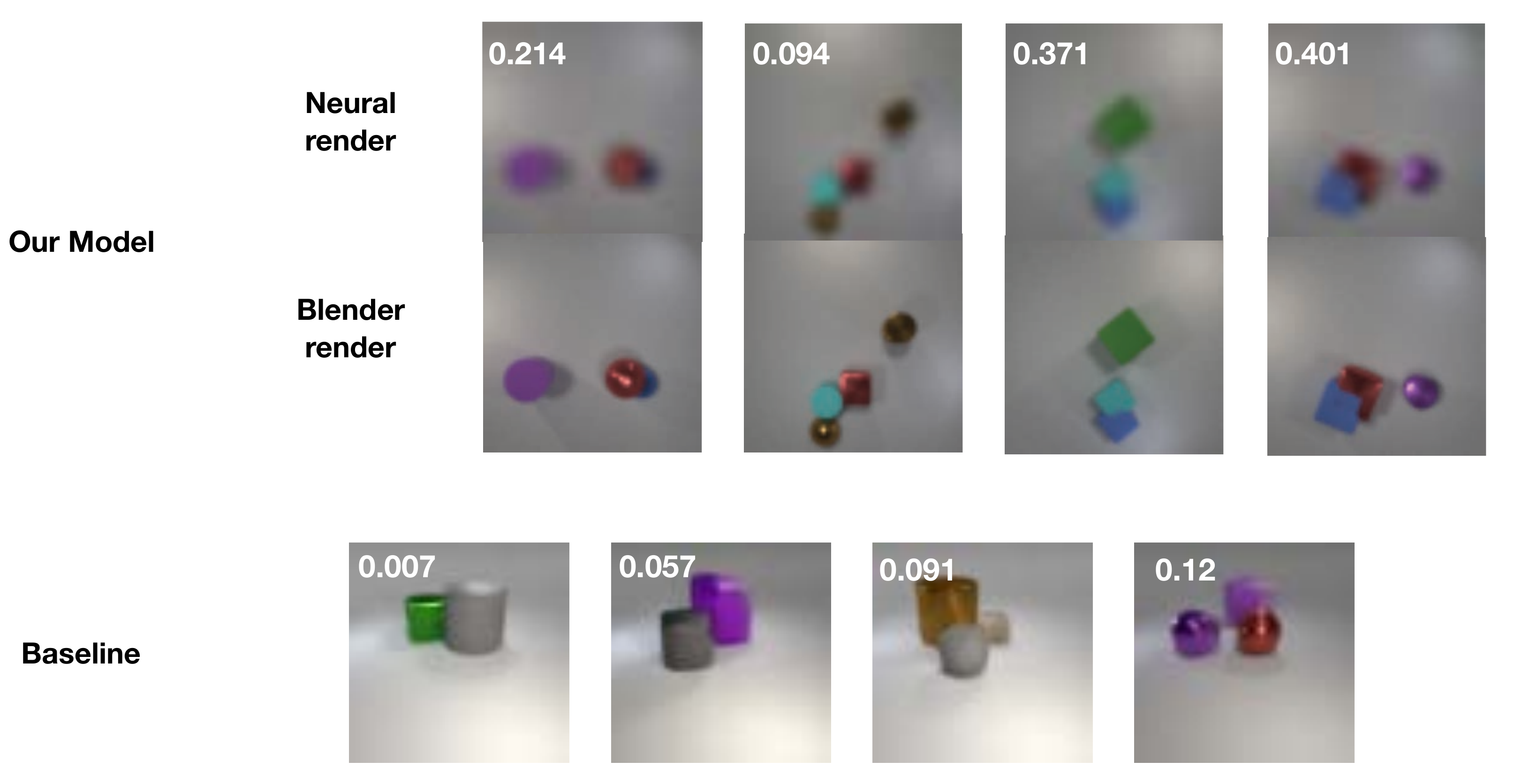}
    \caption{ \textbf{Affordability prediction comparison of our model with the baseline work of \cite{NIPS2018_7658}.} In the top 2 rows, we show the Neural and Blender renderings of our model. Since we reason about the scene in 3D, our model allows checks for expression affordability by computing the 3D intersection-over-union (IoU) scores. In contrast, the bottom row shows the baseline model which operates in 2D latent space and hence cannot differentiate between 2D occlusions and overlapping objects in 3D.
    }
    \label{fig:affordability}
\end{figure*}

 \begin{figure*}[h!]
    \centering
    \includegraphics[width=\textwidth]{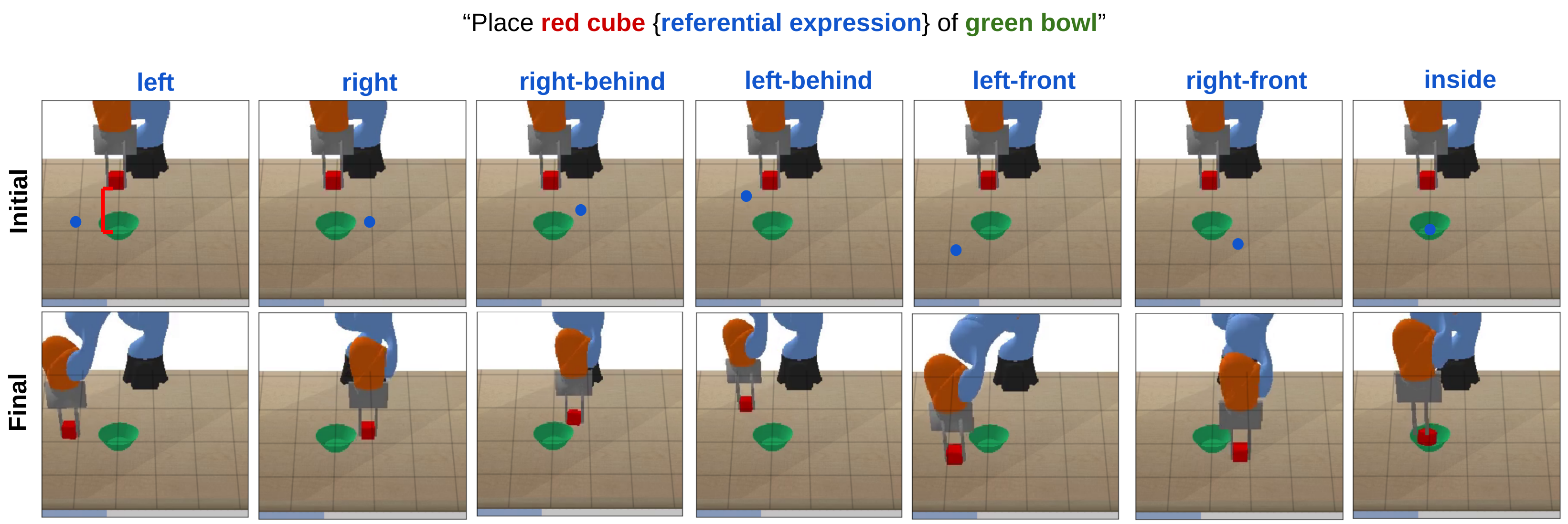}
    \caption{\textbf{Language-guided placement policy learning.} We show the final configurations of the learned policy using different referential expressions for the utterance "Place red cube \{referential expression\} of green bowl." \textit{Top:} Initial robot configuration with the goal position generated by our method indicated as a blue dot. \textit{Bottom:} Final robot configuration. We can see that the robot successfully places the cube with respect to the bowl according to the given referential expression.
    }
    \label{fig:placement1}
\end{figure*}

\section{Additional related work}
The inspiring experiments of Glenberg and Robertson  \cite{Article:00:Glenberg:hallsa}  in 1989 
demonstrated  
that humans can easily judge the plausibility  ---they called it  \textit{affordability}---of natural language utterances, such as  \textit{``he used a newspaper to protect his face from the wind"}, and the implausibility of others, such as \textit{``he used a matchbox to protect his face from the wind"}.  They 
suggest that humans associate words with actual objects in the environment  or prototypes in their imagination that retain perceptual properties---how the objects look---and affordance information  \cite{Gibson1979-GIBTEA}---how the objects can be used.  A natural language utterance is then understood through perceptual and motor \textit{simulations} of explicitly and implicitly mentioned nouns and verbs, in some level of abstraction, that encode such affordances. 
For example, the matchbox is too small to protect a human face from the wind, while a newspaper is both liftable by a human and can effectively cover a face when held appropriately. 
This hypothesis is currently better known as 
simulation semantics \cite{Embodiedmeaning,Feldman06,mentalspatial,embsimmean} and   
has extensive empirical support: reaction times  for visual or motor operations are shorter when  human subjects are shown a related sentence  \cite{b5d80bcf4bfa455cbe66e2f0ffe55d8b,expmental}, and MRI activity is increased in the brain's vision system or motor areas  when human subjects are shown vision- or motor-related linguistic concepts, respectively \cite{FFA,motorareas,saygin}. 
This paper proposes an initial computational model for the simulation semantics hypothesis for the language domain of object spatial arrangements.
}

\end{document}